\def\year{2020}\relax
\documentclass[letterpaper]{article} % DO NOT CHANGE THIS
\usepackage{aaai20}  % DO NOT CHANGE THIS
\usepackage{times}  % DO NOT CHANGE THIS
\usepackage{helvet} % DO NOT CHANGE THIS
\usepackage{courier}  % DO NOT CHANGE THIS
\usepackage[hyphens]{url}  % DO NOT CHANGE THIS
\usepackage{graphicx} % DO NOT CHANGE THIS
\usepackage{graphicx}  % DO NOT CHANGE THIS
\usepackage{graphicx}
\usepackage{amsmath}
\usepackage{booktabs}
\usepackage{url}            % simple URL typesetting
\usepackage{booktabs}       % professional-quality tables
\usepackage{amsfonts}       % blackboard math symbols
\usepackage{nicefrac}       % compact symbols for 1/2, etc.
\usepackage{microtype}      % microtypography
\usepackage{graphicx}
\usepackage{subfigure}
\usepackage{amsmath}
\usepackage{bm}
\usepackage{algorithmic}
\usepackage{algorithm}
 \usepackage{color}

\newcommand{\ignore}[1]{{}}

\def\tvin{{\tt TVIN} }
\def\vin{{\tt VIN} }
\title{Transfer Value Iteration Networks}
\author{Junyi Shen\textsuperscript{\rm 1,2}, Hankz Hankui Zhuo\textsuperscript{\rm 1}\thanks{corresponding author}, Jin Xu\textsuperscript{\rm 2}, Bin Zhong\textsuperscript{\rm 2} \and Sinno Jialin Pan\textsuperscript{\rm 3}\\ % All authors must be in the same font size and format. Use \Large and \textbf to achieve this result when breaking a line
\textsuperscript{\rm 1}School of Data and Computer Science, Sun Yat-Sen University, Guangzhou, China\\
\textsuperscript{\rm 2}Wechat, Tencent Inc., China\\
\textsuperscript{\rm 3}Nanyang Technological University, Singapore\\
vichyshen@tencent.com, zhuohank@mail.sysu.edu.cn, \{jinxxu,harryzhong\}@tencent.com, sinnopan@ntu.edu.sg % email address must be in roman text type, not monospace or sans serif
}
\begin{document}

\maketitle

\begin{abstract}
Value iteration networks (VINs) have been demonstrated to have a good generalization ability for reinforcement learning tasks across similar domains. However, based on our experiments, a policy learned by VINs still fail to generalize well on the domain whose action space and feature space are not identical to those in the domain where it is trained. In this paper, we propose a transfer learning approach on top of VINs, termed Transfer VINs (TVINs), such that a learned policy from a source domain can be generalized to a target domain with only limited training data, even if the source domain and the target domain have domain-specific actions and features. We empirically verify that our proposed TVINs outperform VINs when the source and the target domains have similar but not identical action and feature spaces. Furthermore, we show that the performance improvement is consistent across different environments, maze sizes, dataset sizes as well as different values of hyperparameters such as number of iteration and kernel size.
\end{abstract}
\section{Introduction}
Convolutional neural networks (CNNs) have been applied to reinforcement learning (RL) tasks to learn policies, i.e., a mapping from observations of system states to actions \cite{DBLP:journals/nature/MnihKSRVBGRFOPB15}. As analyzed in \cite{DBLP:conf/nips/TamarLAWT16}, reactive polices learned by conventional CNN-based architectures usually fail to generalize well to previously unseen RL domains even though most of the configurations remain the same to the training domain. To boost the generalization performance, \emph{value iteration networks} (VINs) \cite{DBLP:conf/nips/TamarLAWT16} have been proposed to integrate a planning module into policy learning. VINs have been applied to various application tasks including path planning, e.g., visual navigation \cite{Gupta_2017_CVPR} and the WebNav challenge \cite{DBLP:conf/nips/NogueiraC16}, which requires the agent to navigate the links of a website towards a goal web-page, specified
by a short query. Despite the success of VINs, we observe that the generalizability of VINs is based on an implicit assumption that the feature space and the action space in the unseen domain are as the same as the ones in the seen domain for training the policy. To relax this assumption, in this paper, we propose a transfer learning framework to generate VIN-based policies across different domains even if their action spaces and feature spaces are not identical.
\ignore{When the training sets in these different domains have similar type of features or they share some actions, learning parameters of each VIN network separately will be too expensive and requires much more training data. Knowledge transfer, as is surveye \cite{DBLP:journals/tkde/PanY10}, if done successfully in VINs, would greatly improve the performance of training by reducing the learning expense of parameters. The goal of this paper is to design a transfer learning framework in VINs between different domains to generate VIN-based policies. }

Intuitively, if the target domain has different feature space and action space to the source domain, the VIN-based policy learned from the source domain fails to be used in the target domain directly. A straight-forward solution is to learn a new policy from scratch from the target domain, which is time-consuming. Therefore, it is more desirable to transfer the learned knowledge captured in the source-domain VIN-based policy to the target domain, such that an optimal target policy can be learned with less training data and shorter training time. However, if the source domain and the target domain have totally different feature spaces or action spaces, it is extremely difficult to adapt a learned policy across domains effectively. Therefore, in this work, we assume that 1) the feature spaces between the source and the target domains can be different but are not heterogeneous (e.g., text v.s. images), 2) there is a common subset of actions between the source and the target domains.

We propose the Transfer VIN (\tvin) to transfer the a well-trained VIN-based policy from the source domain to the target domain with limited training data. Specifically, to address the difference between feature spaces and action spaces across domains, in \tvin, we develop two transfer learning modules with respect to the learned reward function and the learned transition function, respectively:
\begin{itemize}
\item We first encode state observations (with different features to the source domain) in the target domain to the same representation of the source domain, such that the reward function transferred from the pre-trained VIN can accurately produce reward images in the target domain.
\item We then leverage the common subset of actions between the source and the target domains to transfer state-action transition information from the source domain to the target domain. Furthermore, we fine-tune the transferred transition function by introducing transfer weights to automatically learn to what degree the transferred actions resemble.
\end{itemize}
By leveraging knowledge transferred via the above transfer learning modules, we further design a new Value Iteration module (VI module) to generate a policy for the target domain. An optimal target-domain policy can be learned by back-propagating the gradient of loss through the whole \tvin in an end-to-end training manner.

To evaluate the effectiveness of our proposed \tvin, we conduct experiments to transfer knowledge between different 2D RL domains, including 2D mazes and Differential Drive \cite{gppn2018}. We evaluate the transfer performance of \tvin with varying environments, maze sizes, dataset sizes and hyperparameters, etc. Extensive experiments empirically show that our proposed \tvin is able to learn a target-domain policy significantly \emph{faster} and reach a \emph{higher} generalization performance, compared with the conventional VIN and another heuristic transfer learning method. %Therefore, representing transfer value iteration networks in this form leads to higher efficiency that accelerates learning process and requires less dataset of the target domain.

\section{Problem Definition}
Let $M$ denote the MDP of some domain, where an optimal policy $\pi$ is expected to be learned. The states, actions, rewards, and transitions in $M$ are denoted by $s \in \mathcal{S}$, $a \in \mathcal{A}$, $R(s, a)$ and $P (s'|s, a)$ respectively.
%where $S$ is a finite set of states, $A$ is a finite set of actions, $P (s'|s, a)$ is defined by $P (s'|s, a) = P(s_{t+1}=s' \mid s_t=s, a_t=a)$ which specifies the probability of state $s'$ at time $t+1$ when applying action $a$ on state $s$ at time $t$, and $R(s, a)$ is the immediate reward received due to action $a$.
Let $\phi(s)$ denote an observation for state $s$. $R$ and $P$ are dependent on the observations as $R = f_R(\phi(s))$ and $P = f_P(\phi(s))$. The functions $f_R$ and $f_P$ are learned jointly in the policy learning process. Given a pre-trained MDP in a source domain, we aim to transfer the learned knowledge including the learned reward function and transition function to the target domain, such that an optimal policy $\pi(a|\phi(s);\theta)$ for the target domain can be learned. Here $\theta$ denotes all the parameters of the \tvin. %Our \tvin policy is represented as a neural network, with the joint parameter $\theta$ denoting the network weights. %For example in a grid-world domain, the input $\phi(s)$ is the full observation vector which consists of the observation image and the state position $s = (i,j)$.

\begin{figure}[t]
 \centering
 \includegraphics[width=0.48\textwidth] {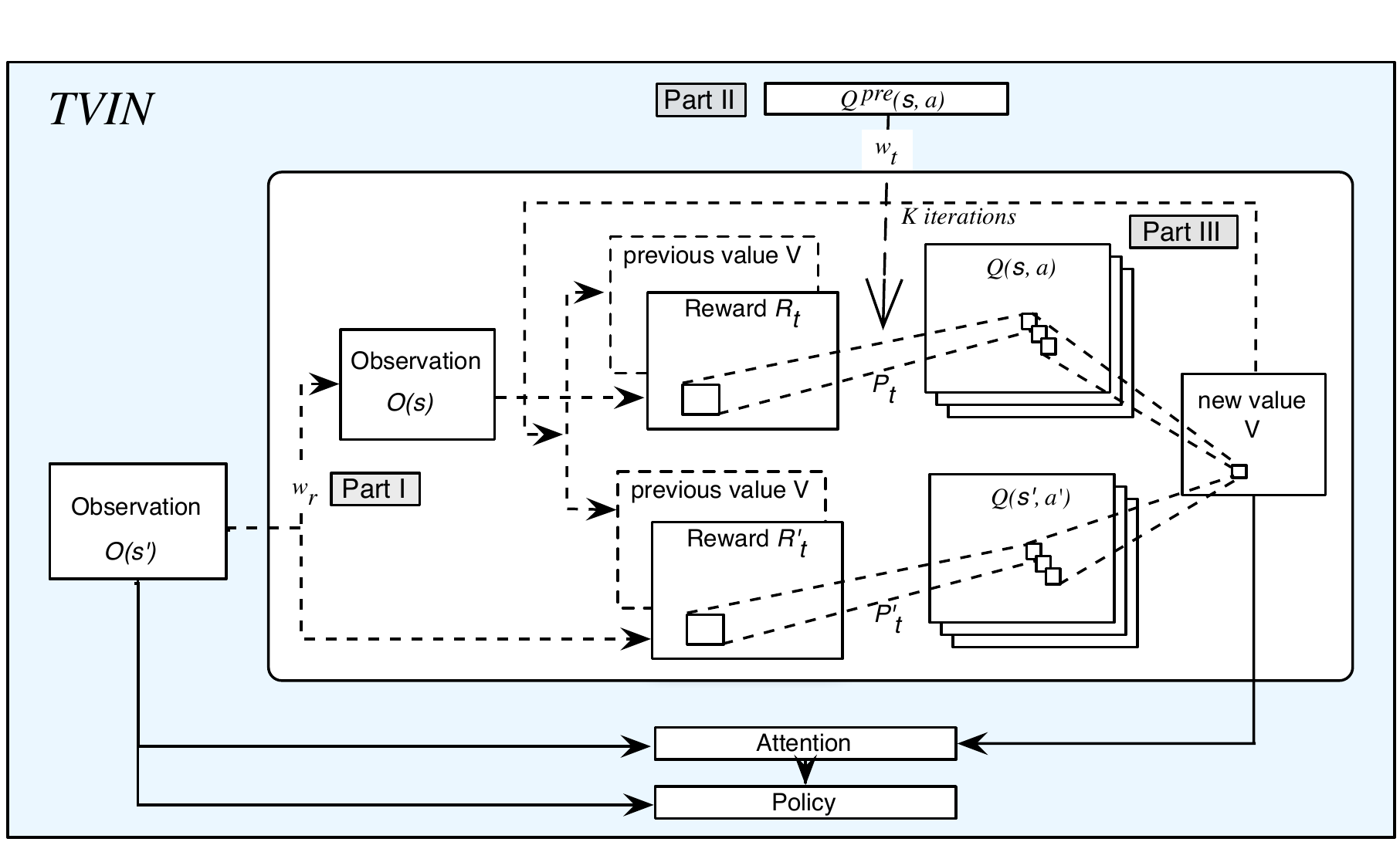}
\caption{\label{1} The Framework of a \tvin}
\end{figure}

\section{Transfer Value Iteration Networks}
In this section, we introduce our proposed \tvin in detail. The overall framework of \tvin is depicted in Figure 1. We develop an encoder to map observations of states in the target domain to the feature representation as the same as the source domain, which is indicated as ``part I'' in the figure. We then transfer the Q-network with respect to \emph{the common subset of actions} from the source domain to the target domain, which is indicated as ``part II'' in the figure. After that, we enrich the Q-network by learning states transition for domain-specific actions of the target domain from scratch. By combining the above two Q-networks, we design a new Value Iteration module (VI module) for the target domain, which is indicated as ``part III''. The planning-integrated \tvin-based policy for the target domain can be trained in an end-to-end manner by back-propagating the gradient through the whole network.

\subsection{Pre-trained VIN}
For \tvin, we suppose that a well-trained source-domain VIN model is given in advance. Basically, a key idea behind many reinforcement learning algorithms is to estimate the action-value function \cite{J2002An}, by using the Bellman equation as an iterative update, $ Q_{i+1}(s,a) = \mathbb{E}_{s,a}\left[r + \gamma \max_{a'}Q_{i}(s',a') |s, a\right] $. The value-iteration algorithm is a popular algorithm for calculating the optimal value function $V^*$ and deriving the correspondingly optimal policy $\pi^*$. In each iteration, $V_{n+1}(s)= \max_a{Q_n(s,a)} \indent \forall{s}$, where $Q_n(s,a) = R(s,a)+ \gamma \sum_{s'} P(s'|s,a)V_n(s')$.
The value function $V_n$ converges to the optimal value function $V^*$ when $n\rightarrow{\infty}$, from which an optimal policy is derived as $\pi^{*}(s) = \arg\max_{a}Q_{\infty}(s, a)$. In a VIN, a VI module implemented by a neural network is used to approximate the value iteration algorithm. Specifically, the VIN first produces a reward image $R$ by $f_R(\phi(s);\theta)$ and inputs $R$ of dimensions $l$, $m$, and $n$ to the VI module. The reward is then fed into a convolutional Q-layer of $\mathcal{A}$ channels followed by a linear activation function: $Q_{a,i',j'} = \sum_{l,i,j} {W_{l,i,j}^{a}R_{l,i'-i,j'-j}}$. Each channel in this layer corresponds to $Q(s, a)$ for a particular action $a$. This layer is then max-pooled along the actions channel to produce the next-iteration value function layer, where ${V}_{i, j} = \max_{{a}} ({Q}(a,i,j))$. The next-iteration value function layer $V$ is then stacked with the reward $R$, and fed back into the convolutional layer and max-pooling layer $K$ times to perform $K$ value iterations. By training the VIN end-to-end in the source domain, we obtain a source-domain VIN and its derived policy for knowledge transfer.
%Pre-trained VINs can learn to plan, and are suitable for predicting outcomes that involve planning-based reasoning, such as policies for reinforcement learning, generalizing better to new, unseen domains.

\subsection{TVIN Algorithm}
The overall algorithm is presented in Algorithm 1. Given the pre-trained VIN in the source domain, the pre-trained reward function $f_R$ is first transferred to produce reward images for the observation $s$ in the target domain (i.e., Step 3). After that the state transition values on the common subset of actions, $f_p^{pre}$, is transferred to the target domain with a learnable weight associated with each action to measure the similarity degree between domains. And the state transition values on new domain-specific actions, $f_p^{new}$, are learned from scratch. All of these state transition values reconstruct a transition function in the target domain, which is further used to compute the Q-function in each iteration for the target domain (i.e., Steps 6 and 7). An attention vector is fed as an input to generate the target policy $\pi_T$ (i.e., Step 11). Finally, the back-propagation algorithm is used to update the parameters of the whole network to learn an optimal target-domain policy (i.e., Step 13). The implementation details of transferring the reward function and the transition function are described in the following sections.

\subsubsection{Reward function transferring} In the source VIN, $f_R(s)$ maps observations of input states to reward images, and pass the reward images to the VI module. For example in the gird-world domain \cite{DBLP:conf/nips/TamarLAWT16}, $f_R$ can map an observation to a high reward at the goal, and negative reward near an obstacle. If we directly adopt the pre-trained $f_R$ from the source domain to the target domain, the reward function may be constrained to the task-specific features due to the diversity of pixel-level inputs. Therefore, for the target domain where the feature space is different from that of the source domain, we propose a feature mapping component to map states from different domains onto the same representation. Specifically, we encode the state observations in the target domain into the same representation as in the source domain by using an autoencoder \cite{DBLP:conf/ijcai/ZhuangCLPH15}. In this way, the reward function transferred from the pre-trained VIN is able to accurately produce a reward image for the target domain before being passed to the new VI module. In particular, we reuse the learned parameters of pre-trained reward function in source domain, and retrain an additional fully-connected layer acting as the feature encoder to output a shared representation for the input states $s$ in the target domain. This feature encoder is trained in an end-to-end manner with the whole \tvin. The new reward function is denoted by $R(s,a) = f_R(s, a;\theta)$, where $\theta$ denotes all the parameters of the whole \tvin.
\begin{algorithm}[t]
\caption{Transfer Value Iteration Algorithm} \label{algorithm:framework}
%{\bf Input:} Observation for state s $\phi(s)$ and the pre-trained $f_R$ and $f_P^{pre}$\\
%{\bf Output:} Planning-based policy $\pi_{T}(a|s;\theta)$

\begin{algorithmic}[1]
\STATE Initialize value function ${V}(s)$ with zeros
%\State $\Delta \leftarrow{0}$
\FOR{epoch = 1, $M$}
	\STATE Set reward $ {R}(s, a) = {f_R(\phi(s), a; \theta)}$
	\FOR{$n$ = 1, $K$}
        		%\FOR{ each $s \in {\mathcal{S}}$}
                \STATE Construct transition functions for each of the states:
                \STATE $
  {P}(s'|s,a) =
  \begin{cases}
    {f_P^{new}(\phi(s), a; \theta)}, &\text{if $a \in{\mathcal{A}_{new}}$}\\
	{\theta_t f_P^{pre}(\phi(s), a)}, &\text{if $a \in{\mathcal{A}_{transfer}}$,}
  \end{cases}
$
            		
                    \STATE $Q_n(s,a; \theta) = R(s,a)+ \gamma \sum_{s'} P(s'|s,a)V_n(s') $
            		\STATE $ {V}_{n+1}(s; \theta) = {\max_{a}{Q_n(s,a; \theta)}}$
        		%\ENDFOR
    	\ENDFOR
    	\STATE Construct optimal $Q$ with $Q^{*}(s, a) = R(s,a)+ \gamma \sum_{s'} P(s'|s,a){V}^{*}(s')$
         	
        \STATE Add attention vector $\psi(s;\theta)$ to the final policy ${\pi}_{T}(a|\psi(s);\theta)$
        \STATE Compute \tvin policy ${\pi}_{T}(a|s;\theta)$ with $\pi^{*}(s) = \arg\max_{a}Q^{*}(s, a) $
    %\State Perform a gradient descent step on $(y-\pi(\phi(s);\theta))^2$
    	\STATE Update $\theta$ by back-propagating the gradient according to \eqref{eq:grad}
\ENDFOR
\end{algorithmic}
\end{algorithm}
\subsubsection{Transition function transferring} To transfer the transition function across domains, we design a new VI module, which performs value iteration by approximating the Bellman-update through a CNN in the target domain. Specifically, the CNN used in the VI module is comprised of stacked convolution and max-pooling layers. The input to each convolution layer is a 3-dimensional signal $X$, typically, an image with $l$ channels and $m\times n$ pixels. Its output $h$ is a $l'$-channel convolution of the image with different kernels: $ h_{l',i',j'} = \sigma{\left( \sum_{l,i,j} {W_{l,i,j}^{l'}X_{l,i'-i,j'-j}} \right)}$, where $\sigma$ is an activation function. A max-pooling layer then down-samples the image by selecting the maximum value among some dimension.
%CNNs are typically trained using stochastic gradient descent (SGD), with back-propagation for computing gradients.
In this sense, each iteration in our new VI module can be approximately regarded as passing the reward $R$ as well as the previous value function $V_n$ through a convolution layer and max-pooling layer. As mentioned in \cite{DBLP:conf/nips/TamarLAWT16}, each channel in the convolution layer corresponds to the Q-function for a specific action, and convolution kernel weights correspond to the discounted transition probabilities. Thus, we leverage the state transition values regarding the common subset of actions from the pre-trained model as a bridge of transition functions between the source domain and the target domain. At the high level, the new VI module divides the channels in the convolution layer into two parts: one corresponds to the Q-function for the common subset of actions and the other is for the new actions in the target domain.

For common actions between domains, some of them perform more similarly on both domains, while others may perform less similarly. To model the degree of similarity of actions between domains, we propose to add a learnable weight $\theta_t$ for each common action. The fine-tuned transition function in the target domain is defined by ${P}(s'|s,a) = \theta_t f_P^{pre}(\phi(s), a)$, if $a\in{\mathcal{A}_{transfer}}$, and ${P}(s'|s,a) = f_P^{new}(\phi(s), a)$, if $a\in{\mathcal{A}_{new}}$, where $\mathcal{A}_{transfer}$ is the common subset of actions, $f_P^{pre}$ is adopted from the pre-trained transition function, $\mathcal{A}_{new}$ is the subset of domain-specific actions in the target domain, and $f_P^{new}$ is learned from scratch with target domain data.
When back-propagating the gradient through the \tvin in the target domain, we fix $f_P^{pre}$ and only learn $\theta_t$ and $f_P^{new}$. To sum up, convolution kernel weights corresponding to the discounted transition probabilities in \tvin are computed based on two different cases:
\begin{equation}
  W^{a} =\left\{
   \begin{array}{rcl}
\theta_t W_{pre}^{a} & a\in{\mathcal{A}_{transfer}};\\
W_{new}^{a} & a\in{\mathcal{A}_{new}},
\end{array}
   \right.
\end{equation}
where $W_{pre}^{a}$ stands for the pre-trained convolution kernel parameters corresponding to the discounted transition probabilities for the transferred (common) actions, $\theta_t$ stands for the transfer weight, and $W_{new}^{a}$ stands for the new convolution kernel parameters corresponding to the discounted transition probabilities for new actions in target domain. The value function ${V}$ is stacked with the reward ${R}$, and they are fed back into the convolutional layer, where each channel corresponds to the Q-function for a specific action. The convolution operation in new VI module is formulated as
\begin{equation}
 Q_{a,i',j'} = \sum_{l,i,j} {W_{l,i,j}^{a}R_{l,i'-i,j'-j}}+\sum_{i,j} {W_{l+1,i,j}^{a} {V}_{i'-i,j'-j}},
\end{equation}
where $i$, $j$ stand for the input state $s = (i, j)$, $R$ is the reward and $V$ is the value function in each iteration. These convolutional channels in both two parts are then max-pooled along all channels to produce the next-iteration value function layer $V$ with $ {V}_{n+1}(s; \theta) = {\max_{a}{Q_n(s,a; \theta)}}$. Performing value iteration for $K$ times in this form, the new VI module outputs the approximate optimal value function ${V}^{*} = {V}_{K}$. The value iteration module in \tvin has an effective depth of $K$, which is larger than the depth of the well-known Deep Q-Network \cite{DBLP:journals/nature/MnihKSRVBGRFOPB15}. To reduce parameters for training process, we share the weights in the $K$ recurrent layers in the \tvin.

After learning the internal transfer VI module which is independent to observations, we generate a policy for the input state $s$ according to
$
{\pi}^{*}(s) = \arg\max_{a}R(s,a)+ \gamma \sum_{s'} P(s'|s,a){V}^{*}(s')
$.
Note that the transition $\sum_{s'} P(s'|s,a){V}^{*}(s')$ only depends on a subset of the optimal value function ${V}^{*}$, if the states have a topology with local transition dynamics such as the grid-world application. Thus, we suppose that a local subset of $s$ is sufficient for extracting information about the optimal \tvin plan.

Motivated by the wide use of attention mechanism \cite{DBLP:conf/icml/XuBKCCSZB15} to improve learning performance by reducing the effective number of network parameters during training, in \tvin, we introduce an attention moduel to select the value of the current state after $K$ iterations of value iteration. Intuitively, for a given label prediction (action), only a subset of the input features (value function) is relevant. The attention module can be represented by a parameterized function to output an attention modulated vector $\psi(s;\theta)$ for the input state $s$. And this vector is added as additional features to the \tvin to generate the final policy $\pi_T(\psi(s);\theta)$. By back-propagating through the whole network in an end-to-end manner, we update the joint parameters $\theta$ and learn the planning-based \tvin policy for the target domain.

\subsection{Updating Parameters}
By specifying the forms of the reward function $f_R$, the transition function $f_P$, and the attention function, and denoting the parameters of the whole \tvin by $\theta$, we define the policy objective over the \tvin as the cross-entropy loss function between the expert policy and the current policy derived by \tvin. The \tvin can be trained by minimizing the loss function $\mathcal{L(\theta)}$,
\begin{equation}
\mathcal{L}(\theta) = \sum_{a\in{\mathcal{A}}}{\pi_E(a|s)\log{\pi_T(a|s;\theta)}},
\end{equation}
where $\pi_{T}(a|s;\theta)$ is the \tvin policy parameterized by $\theta$, and $\pi_E(a|s)$ is the expert policy for training data. To acquire training data, we can sample the expert to generate the trajectories used in the loss. In contrast to the deep reinforcement learning objective \cite{DBLP:journals/nature/MnihKSRVBGRFOPB15} which recursively relies on itself as a target value, we use imitation learning (IL) \cite{DBLP:journals/ral/GiustiGCHRFFFSC16}, which uses a stable training signal generated by an expert to guide the transfer network. Learning the \tvin policy then becomes an instance of supervised learning.

We consider the updates that optimize the policy parameter $\theta$ of the state representation, the reward function, and the new VI model. We update the $\theta$ towards the expert outcome. The gradient of the loss function with respect to the weights can be computed via
\begin{equation}\label{eq:grad}
\nabla_{\theta}\mathcal{L}(\theta) = \sum_{a\in{\mathcal{A}}} \frac{\pi_E(a|s)}{\pi_T(a|s;\theta)}\nabla_{\theta}\pi_T(a|s;\theta).
\end{equation}
We use the above gradient to update parameters by stochastic gradient descent (SGD) \cite{Boyd:2004:CO:993483}. In summary, the joint parameters $\theta$ of are updated to make the planning-integrated \tvin-based policy $\pi_T$ more close to the expert policy $\pi_E$.

\ignore{
\begin{equation}
\frac{\partial \mathcal{L}}{\partial \theta} = \sum_{a\in{\mathcal{A}}} \frac{\pi_E(a|s)}{\pi_T(a|s;\theta)}\frac{\partial \pi}{\partial \theta}
\end{equation}

\begin{equation}
\mathcal{L}(\theta) = \mathbb{E}_{s,a}\left[(y-\pi(\phi(s);\theta))^2\right]
\end{equation}

\begin{equation}
\frac{\partial \mathcal{L}}{\partial \theta} = (y-\pi(s;\theta)) \frac{\partial \pi}{\partial \theta}
\end{equation}

\begin{equation}
\frac{\partial \mathcal{L}}{\partial \theta} = (\bm{y}-\bm{g}) \frac{\partial \bm{g}}{\partial \bm{\theta}}
\end{equation}
}
\ignore{
 \begin{figure*}[t]
 \centering
 \begin{minipage}[t]{0.5\textwidth}
\centering
\includegraphics[width=.45\textwidth]{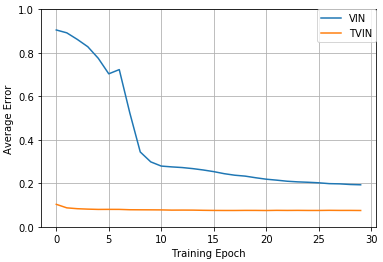}
\includegraphics[width=.45\textwidth]{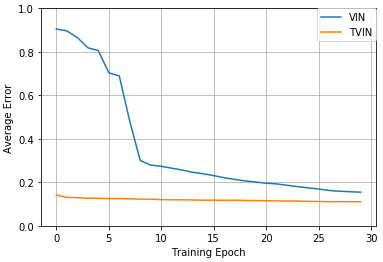}
\end{minipage}%
\begin{minipage}[t]{0.5\textwidth}
\centering
\includegraphics[width=.4\textwidth]{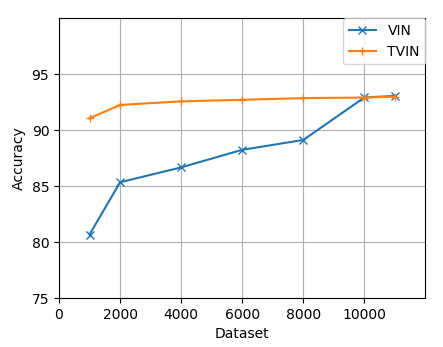}
\includegraphics[width=.4\textwidth]{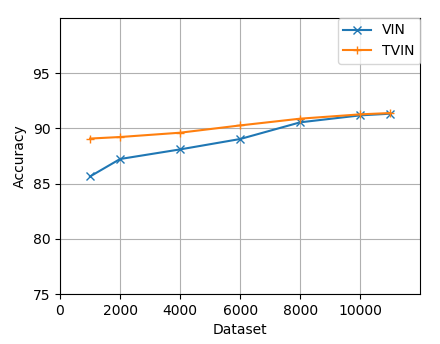}
\end{minipage}

 \caption{\label{3} Transfer from NEWS-9 to Moore-15 with NEWS-9 pre-trained with 1k training data. Left: Training process compared with VIN. A: domains of 30\% obstacles. B: domains of 50\% obstacles. Right: \tvin performance with varying training data. C: domains of 30\% obstacles.D: domains of 50\% obstacles.}
 \end{figure*}
 }
\begin{table*}[]
\resizebox{1\textwidth}{18.6mm}{
\begin{tabular}{ccccccccccc}
\toprule[1pt]

\multicolumn{2}{c|}{Source}                       & \multicolumn{3}{c|}{NEWS-9}                                                                                                                   & \multicolumn{3}{c|}{NEWS-15}                                                                                                                  & \multicolumn{3}{c}{NEWS-28}                                                                                             \\ \hline
\multicolumn{2}{c|}{Target}                       & \multicolumn{1}{c|}{Moore-9}                  & \multicolumn{1}{c|}{Moore-15}                 & \multicolumn{1}{c|}{Moore-28}                 & \multicolumn{1}{c|}{Moore-9}                  & \multicolumn{1}{c|}{Moore-15}                 & \multicolumn{1}{c|}{Moore-28}                 & \multicolumn{1}{c|}{Moore-9}                  & \multicolumn{1}{c|}{Moore-15}                 & Moore-28                 \\ \hline
N                    & \multicolumn{1}{c|}{Model} & \multicolumn{1}{c|}{\%Opt    \%Suc}           & \multicolumn{1}{c|}{\%Opt    \%Suc}           & \multicolumn{1}{c|}{\%Opt    \%Suc}           & \multicolumn{1}{c|}{\%Opt    \%Suc}           & \multicolumn{1}{c|}{\%Opt    \%Suc}           & \multicolumn{1}{c|}{\%Opt    \%Suc}           & \multicolumn{1}{c|}{\%Opt    \%Suc}           & \multicolumn{1}{c|}{\%Opt    \%Suc}           & \%Opt    \%Suc           \\

\midrule[0.5pt]

1k                   & \multicolumn{1}{c|}{VIN}   & \multicolumn{1}{c|}{84.2\quad     87.7}          & \multicolumn{1}{c|}{77.3\quad     81.7}          & \multicolumn{1}{c|}{56.2\quad     65.8}          & \multicolumn{1}{c|}{84.2\quad     87.7}          & \multicolumn{1}{c|}{77.3\quad     81.7}          & \multicolumn{1}{c|}{56.2\quad     65.8}          & \multicolumn{1}{c|}{84.2\quad     87.7}          & \multicolumn{1}{c|}{77.3\quad     81.7}          & 56.2\quad     65.8          \\
1k                   & \multicolumn{1}{c|}{TVIN}  & \multicolumn{1}{c|}{\textbf{89.8\quad     94.2}} & \multicolumn{1}{c|}{\textbf{88.3\quad     91.0}} & \multicolumn{1}{c|}{\textbf{66.7 \quad    74.7}} & \multicolumn{1}{c|}{\textbf{94.6\quad     96.6}} & \multicolumn{1}{c|}{\textbf{90.1\quad     92.8}} & \multicolumn{1}{c|}{\textbf{66.1\quad     75.3}} & \multicolumn{1}{c|}{\textbf{94.3\quad     95.8}} & \multicolumn{1}{c|}{\textbf{86.4\quad     89.1}} & \textbf{62.1\quad     71.1} \\
\specialrule{0em}{1pt}{1pt}
\hline
\specialrule{0em}{1pt}{1pt}
5k                   & \multicolumn{1}{c|}{VIN}   & \multicolumn{1}{c|}{90.5\quad     92.5}          & \multicolumn{1}{c|}{86.7\quad     88.7}          & \multicolumn{1}{c|}{64.3\quad     72.9}          & \multicolumn{1}{c|}{90.5\quad     92.5}          & \multicolumn{1}{c|}{86.7\quad     88.7}          & \multicolumn{1}{c|}{64.3\quad     72.9}          & \multicolumn{1}{c|}{90.5\quad     92.5}          & \multicolumn{1}{c|}{86.7\quad     88.7}          & 64.3\quad     72.9          \\
5k                   & \multicolumn{1}{c|}{TVIN}  & \multicolumn{1}{c|}{\textbf{97.0\quad     98.0}} & \multicolumn{1}{c|}{\textbf{93.8\quad     94.9}} & \multicolumn{1}{c|}{\textbf{80.4\quad     86.3}} & \multicolumn{1}{c|}{\textbf{97.1\quad     97.2}} & \multicolumn{1}{c|}{\textbf{95.2\quad     96.0}} & \multicolumn{1}{c|}{\textbf{73.4\quad     84.3}} & \multicolumn{1}{c|}{\textbf{97.8\quad     98.2}} & \multicolumn{1}{c|}{\textbf{91.1\quad    92.6}} & \textbf{76.2\quad     84.3} \\
\specialrule{0em}{1pt}{1pt}
\hline
\specialrule{0em}{1pt}{1pt}
10k                  & \multicolumn{1}{c|}{VIN}   & \multicolumn{1}{c|}{86.2\quad     88.0}          & \multicolumn{1}{c|}{91.1\quad     92.3}          & \multicolumn{1}{c|}{60.8\quad     68.0}          & \multicolumn{1}{c|}{86.2\quad     88.0}          & \multicolumn{1}{c|}{91.1\quad     92.3}          & \multicolumn{1}{c|}{60.8\quad     68.0}          & \multicolumn{1}{c|}{86.2\quad     88.0}          & \multicolumn{1}{c|}{91.1\quad     92.3}          & 60.8\quad     68.0          \\
10k                  & \multicolumn{1}{c|}{TVIN}  & \multicolumn{1}{c|}{\textbf{97.6\quad     97.8}} & \multicolumn{1}{c|}{\textbf{95.4\quad     96.2}} & \multicolumn{1}{c|}{\textbf{83.1\quad     88.3}} & \multicolumn{1}{c|}{\textbf{97.4\quad     97.5}} & \multicolumn{1}{c|}{\textbf{96.2\quad     96.7}} & \multicolumn{1}{c|}{\textbf{87.8\quad     91.8}} & \multicolumn{1}{c|}{\textbf{96.6\quad     96.8}} & \multicolumn{1}{c|}{\textbf{92.5\quad     93.7}} & \textbf{78.7\quad     84.0} \\

\bottomrule[1pt]

\end{tabular}}
\caption{Transfer from NEWS to Moore with varying dataset sizes N and maze sizes M. }
\end{table*}
\begin{table*}[]
\centering
\resizebox{1\textwidth}{18.6mm}{
\begin{tabular}{ccccccccccc}
\toprule[1pt]
\multicolumn{2}{c|}{Source}                       & \multicolumn{3}{c|}{Moore-9}                                                                                                               & \multicolumn{3}{c|}{Moore-15}                                                                                                              & \multicolumn{3}{c}{Moore-28} \\ \hline
\multicolumn{2}{c|}{Target}                       & \multicolumn{1}{c|}{NEWS-9}                  & \multicolumn{1}{c|}{NEWS-15}                 & \multicolumn{1}{c|}{NEWS-28}                 & \multicolumn{1}{c|}{NEWS-9}                  & \multicolumn{1}{c|}{NEWS-15}                 & \multicolumn{1}{c|}{NEWS-28}                 & \multicolumn{1}{c|}{NEWS-9}                  & \multicolumn{1}{c|}{NEWS-15}                 & NEWS-28                 \\ \hline
N                    & \multicolumn{1}{c|}{Model} & \multicolumn{1}{c|}{\%Opt    \%Suc}          & \multicolumn{1}{c|}{\%Opt    \%Suc}          & \multicolumn{1}{c|}{\%Opt    \%Suc}          & \multicolumn{1}{c|}{\%Opt    \%Suc}          & \multicolumn{1}{c|}{\%Opt    \%Suc}          & \multicolumn{1}{c|}{\%Opt    \%Suc}          & \multicolumn{1}{c|}{\%Opt    \%Suc}          & \multicolumn{1}{c|}{\%Opt    \%Suc}          & \%Opt    \%Suc          \\

\midrule[0.5pt]

1k                   & \multicolumn{1}{c|}{VIN}   & \multicolumn{1}{c|}{77.8\quad      81.0}          & \multicolumn{1}{c|}{69.3\quad      71.1}          & \multicolumn{1}{c|}{45.6\quad      51.9}          & \multicolumn{1}{c|}{77.8\quad      81.0}          & \multicolumn{1}{c|}{69.3\quad      71.1}          & \multicolumn{1}{c|}{45.6\quad      51.9}          & \multicolumn{1}{c|}{77.8\quad      81.0}          & \multicolumn{1}{c|}{69.3\quad      71.1}          & 45.6\quad      51.9          \\
1k                   & \multicolumn{1}{c|}{TVIN}  & \multicolumn{1}{c|}{\textbf{94.7\quad      94.8}} & \multicolumn{1}{c|}{\textbf{85.5\quad      86.8}} & \multicolumn{1}{c|}{\textbf{69.1\quad      71.6}} & \multicolumn{1}{c|}{\textbf{94.8\quad      94.9}} & \multicolumn{1}{c|}{\textbf{96.3\quad      96.4}} & \multicolumn{1}{c|}{\textbf{89.2\quad89.4}} & \multicolumn{1}{c|}{\textbf{82.0\quad      84.0}} & \multicolumn{1}{c|}{\textbf{73.1\quad75.0}} & \textbf{64.0\quad  67.7} \\
\specialrule{0em}{1pt}{1pt}
\hline
\specialrule{0em}{1pt}{1pt}
5k                   & \multicolumn{1}{c|}{VIN}   & \multicolumn{1}{c|}{79.8\quad      81.9}          & \multicolumn{1}{c|}{70.7\quad      73.5}          & \multicolumn{1}{c|}{57.8\quad      60.9}          & \multicolumn{1}{c|}{79.8\quad      81.9}          & \multicolumn{1}{c|}{70.7\quad      73.5}          & \multicolumn{1}{c|}{57.8\quad      60.9}          & \multicolumn{1}{c|}{79.8\quad      81.9}          & \multicolumn{1}{c|}{70.7\quad      73.5}          & 57.8\quad      60.9          \\
5k                   & \multicolumn{1}{c|}{TVIN}  & \multicolumn{1}{c|}{\textbf{95.0\quad      95.0}} & \multicolumn{1}{c|}{\textbf{88.6\quad      89.4}} & \multicolumn{1}{c|}{\textbf{75.1\quad      77.8}} & \multicolumn{1}{c|}{\textbf{97.1\quad97.1}} & \multicolumn{1}{c|}{\textbf{96.5\quad      96.6}} & \multicolumn{1}{c|}{\textbf{93.0\quad      93.1}} & \multicolumn{1}{c|}{\textbf{85.1\quad      86.7}} & \multicolumn{1}{c|}{\textbf{77.3\quad80.3}} & \textbf{65.1\quad      68.1} \\
\specialrule{0em}{1pt}{1pt}
\hline
\specialrule{0em}{1pt}{1pt}
10k                  & \multicolumn{1}{c|}{VIN}   & \multicolumn{1}{c|}{87.1\quad      88.4}          & \multicolumn{1}{c|}{88.1\quad      88.4}          & \multicolumn{1}{c|}{58.4 \quad     61.5}          & \multicolumn{1}{c|}{87.1\quad     88.4}          & \multicolumn{1}{c|}{88.1\quad    88.4}          & \multicolumn{1}{c|}{58.4\quad     61.5}          & \multicolumn{1}{c|}{87.1\quad   88.4}          & \multicolumn{1}{c|}{88.1\quad    88.4}          & 58.4\quad     61.5          \\
10k                  & \multicolumn{1}{c|}{TVIN}  & \multicolumn{1}{c|}{\textbf{96.6\quad      96.6}} & \multicolumn{1}{c|}{\textbf{89.3\quad      90.0}} & \multicolumn{1}{c|}{\textbf{80.1\quad      82.2}} & \multicolumn{1}{c|}{\textbf{97.4\quad      97.4}} & \multicolumn{1}{c|}{\textbf{97.0\quad      96.9}} & \multicolumn{1}{c|}{\textbf{94.4\quad      94.5}} & \multicolumn{1}{c|}{\textbf{91.7\quad      92.5}} & \multicolumn{1}{c|}{\textbf{88.7\quad      89.6}} & \textbf{68.4\quad      72.9} \\
\bottomrule[1pt]

\end{tabular}}
\caption{Transfer from Moore to NEWS with varying dataset sizes N and maze sizes M.}
\end{table*}
\section{Experiments}
\subsection{Datasets and Criteria}
\paragraph{Dataset} The RL task domains for our experiments are synthetic 2D maps with randomly placed obstacles, in which observations include positions of agents, goal positions and the map configurations. Specifically, we use three different 2D environments similar to the GPPN experiments in \cite{gppn2018} : the NEWS, the Moore and the Differential Drive. In NEWS, the agent can move \textbf{\{East, West, North, South\}}; in Differential Drive, the agent can move forward along its current orientation, or turn left/right by 90 degrees. The action space is \textbf{\{Move forward, Turn left, Turn right\}}; in Moore, the agent can move to any of the eight cells in its neighborhood. The action space of Moore is \textbf{\{East, West, North, South, Northeast, Northwest, Southeast, Southwest\}}.  When considering knowledge transfer in the following experiments, we give the pairs of possible similar actions between different domains. Between NEWS and Differential Drive, the similar pairs are \{(North, Move forward), (East, Turn left), (West, Turn right)\}. Between NEWS and Moore, the similar pairs are \{(East, East), (West, West), (North, North), (South, South)\}.

In the experiments on the above three domains, the state vectors given as input to the models consist of the maps and the goal location. In NEWS and Moore, the target is an x-y coordinate. Similar to the experimental setup in \cite{DBLP:conf/nips/TamarLAWT16}, we produce a $(2\times m\times m)$-sized observation image for each state $s = (i,j)$ in each trajectory, where $m$ is the maze size. The first channel of the image encodes the obstacle configuration (1 for obstacle, 0 otherwise), while the second channel encodes the goal position (1 at the goal, 0 otherwise). The full state observation vector consists of the observation image and the state $s = (i,j)$. While in Differential Drive, the goal location contains an orientation along with the x-y coordinate. Consequently, the dimension of the goal map given as input to the models is $4*m*m$ in Differential Drive. In addition, for each state, we produce a ground-truth label encoding the action that an optimal shortest-path policy would take in that state. Experimentally, our ground-truth label is created with a maze generation process that uses depth-first search with the recursive back-tracker algorithm \cite{Cormen:2009:IAT:1614191}.
%\vspace{-1em}
\paragraph{Criteria}
In the following experiments, we empirically compare \tvin and \vin using two metrics referred to \cite{gppn2018} : \%Optimal (\%Opt) is the percentage of states whose predicted paths under the policy estimated by the model has optimal length. \%Opt is denoted by:
$$
\%Opt = \frac{Num(a_i = a_i^*)}{S_{test}},
$$
where $S_{test}$ represents the total number of states in test set, $a_i^*$ represents the optimal action for state $s_i$, and $a_i$ is the action prediction generated by models for $s_i$. The second metric \%Success (\%Suc) is the percentage of states whose predicted paths under the policy estimated by the model reach the goal state. A trajectory is said to succeed if it reached the goal without hitting obstacles. Let $N_{test}$ denote the total number of test trajectories, $\mathcal{T}_{goal}$ represents the goal state of trajectory $\mathcal{T}$ and $\mathcal{T}_{end}$ represents the end state of the trajectory predicted by the models. Then \%Suc can be denoted by:
$$
\%Suc = \frac{Num(\mathcal{T}_{end} = \mathcal{T}_{goal})}{N_{test}}.
$$
\begin{figure}[t]
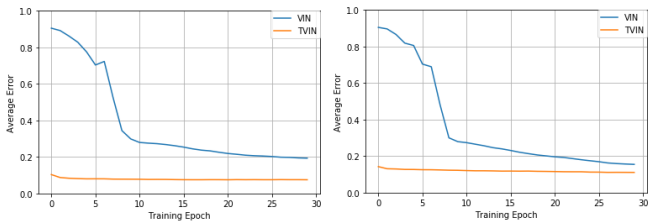

\centering
\begin{minipage}[t]{0.5\textwidth}
\centering
\includegraphics[width=.48\textwidth]{ex_1.png}
\includegraphics[width=.48\textwidth]{ex_2.png}
\end{minipage}%
 \caption{\label{3} Training process on Moore-15 with 1k training data transferred from NEWS-9 compared with VIN. Left: domains of 30\% obstacles. Right: domains of 50\% obstacles.}
 \end{figure}
  \begin{figure}[t]
\centering
\begin{minipage}[t]{0.5\textwidth}
\centering
\includegraphics[width=.48\textwidth]{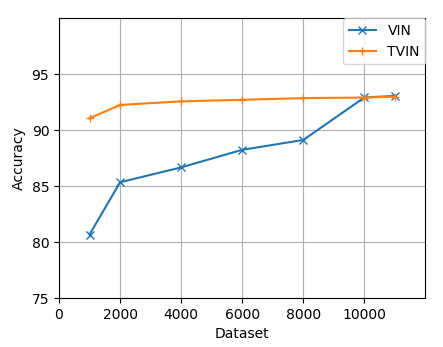}
\includegraphics[width=.48\textwidth]{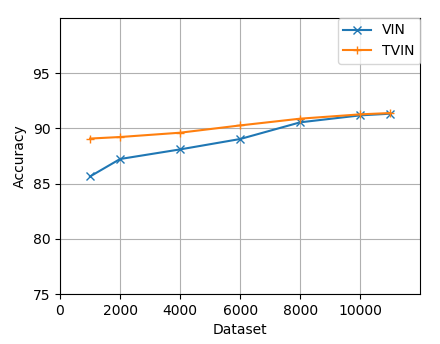}
\end{minipage}%
 \caption{\label{3} Prediction accuracy on Moore-15 with varying training data transferred from NEWS-9 compared with VIN. Left: domains of 30\% obstacles. Right: domains of 50\% obstacles.}
 \end{figure}
\subsection{Experimental Results}
Our experiments attempt to transfer policies between 2D domains with different environments and maze sizes. We evaluate our \tvin approach in the following aspects:
\begin{enumerate}
\item We first evaluate \tvin between different domains, including transfer from NEWS to Moore, transfer from Moore to NEWS and transfer from Differential Drive to NEWS. Additionally we vary the maze sizes in each domains, dataset sizes in the target domains, etc., to see the performance of \tvin when only limited training data is available.
\item We then evaluate \tvin approach on hyperparameter sensitivity, including the iteration count $K$ and the kernel size $F$. Experiments show that \tvin can indeed perform better than single VIN and does not rely on the setting of these hyperparameters.
\item We finally evaluate \tvin by varying the amount of pre-trained knowledge transferred from the source domain, which is characterized by the number of transferable actions between source and target domains. We aim to see the impact of the amount of transferred knowledge.
\end{enumerate}

In 2D domains, an optimal policy can be calculated by exact value iteration algorithm. And the pre-trained VIN represented by a neural network has been proved to learn planning results. However, for these different tasks of similar complexity and sharing similar actions, \tvin can greatly accelerate training process as well as improving the performance of training by leveraging learned knowledge and by reducing the learning expense of parameters.
\begin{table*}[t]

\resizebox{1\textwidth}{18.6mm}{
\begin{tabular}{ccccccccccc}
\toprule[1pt]

\multicolumn{2}{c|}{Source}                       & \multicolumn{3}{c|}{Drive-9}                                                                                                                   & \multicolumn{3}{c|}{Drive-15}                                                                                                                  & \multicolumn{3}{c}{Drive-28}                                                                                             \\\hline

\multicolumn{2}{c|}{Target}                       & \multicolumn{1}{c|}{NEWS-9}                  & \multicolumn{1}{c|}{NEWS-15}                 & \multicolumn{1}{c|}{NEWS-28}                 & \multicolumn{1}{c|}{NEWS-9}                  & \multicolumn{1}{c|}{NEWS-15}                 & \multicolumn{1}{c|}{NEWS-28}                 & \multicolumn{1}{c|}{NEWS-9}                  & \multicolumn{1}{c|}{NEWS-15}                 & NEWS-28                 \\\hline

N                    & \multicolumn{1}{c|}{Model} & \multicolumn{1}{c|}{\%Opt    \%Suc}           & \multicolumn{1}{c|}{\%Opt    \%Suc}           & \multicolumn{1}{c|}{\%Opt    \%Suc}           & \multicolumn{1}{c|}{\%Opt    \%Suc}           & \multicolumn{1}{c|}{\%Opt    \%Suc}           & \multicolumn{1}{c|}{\%Opt    \%Suc}           & \multicolumn{1}{c|}{\%Opt    \%Suc}           & \multicolumn{1}{c|}{\%Opt    \%Suc}           & \%Opt    \%Suc           \\

\midrule[0.5pt]

1k    &\multicolumn{1}{c|}{VIN}   & \multicolumn{1}{c|}{77.8\quad81.0} & \multicolumn{1}{c|}{69.3\quad71.1} & \multicolumn{1}{c|}{45.6\quad51.9} & \multicolumn{1}{c|}{77.8\quad81.0} & \multicolumn{1}{c|}{69.3\quad71.1} & \multicolumn{1}{c|}{45.6\quad51.9} & \multicolumn{1}{c|}{77.8\quad81.0} & \multicolumn{1}{c|}{69.3\quad71.1}  & \multicolumn{1}{c}{45.6\quad51.9} \\

1k     & \multicolumn{1}{c|}{TVIN}  & \multicolumn{1}{c|}{ \textbf{86.7\quad88.1}} & \multicolumn{1}{c|}{ \textbf{70.2\quad72.2}} & \multicolumn{1}{c|}{ \textbf{49.2\quad52.6}} & \multicolumn{1}{c|}{ \textbf{80.0\quad81.4}} & \multicolumn{1}{c|}{ \textbf{83.4\quad84.6}} & \multicolumn{1}{c|}{ \textbf{63.9\quad68.7}} & \multicolumn{1}{c|}{ \textbf{78.3\quad81.3}} & \multicolumn{1}{c|}{ \textbf{72.8\quad74.8}}  & \multicolumn{1}{c}{ \textbf{57.5\quad59.8}} \\
\specialrule{0em}{1pt}{1pt}
\hline
\specialrule{0em}{1pt}{1pt}

5k    & \multicolumn{1}{c|}{VIN}   & \multicolumn{1}{c|}{79.8\quad81.9} & \multicolumn{1}{c|}{70.7\quad73.5} & \multicolumn{1}{c|}{57.8\quad60.9} & \multicolumn{1}{c|}{79.8\quad81.9} & \multicolumn{1}{c|}{70.7\quad73.5} & \multicolumn{1}{c|}{57.8\quad60.9} & \multicolumn{1}{c|}{79.8\quad81.9} & \multicolumn{1}{c|}{70.7\quad73.5}  & \multicolumn{1}{c}{57.8\quad60.9} \\

5k    & \multicolumn{1}{c|}{TVIN}  & \multicolumn{1}{c|}{ \textbf{88.0\quad88.8}} & \multicolumn{1}{c|}{ \textbf{83.7\quad86.0}} & \multicolumn{1}{c|}{ \textbf{84.1\quad84.9}} & \multicolumn{1}{c|}{ \textbf{86.0\quad86.8}} & \multicolumn{1}{c|}{ \textbf{93.4\quad93.6}} & \multicolumn{1}{c|}{ \textbf{91.9\quad92.1}} & \multicolumn{1}{c|}{ \textbf{81.9\quad84.4}} & \multicolumn{1}{c|}{ \textbf{85.2\quad85.1}}  & \multicolumn{1}{c}{ \textbf{78.8\quad80.5}}  \\
\specialrule{0em}{1pt}{1pt}
\hline
\specialrule{0em}{1pt}{1pt}

10k    & \multicolumn{1}{c|}{VIN}   & \multicolumn{1}{c|}{87.1\quad88.4} & \multicolumn{1}{c|}{88.1\quad88.4} & \multicolumn{1}{c|}{58.4\quad61.5} & \multicolumn{1}{c|}{87.1\quad88.4} & \multicolumn{1}{c|}{88.1\quad88.4} & \multicolumn{1}{c|}{58.4\quad61.5} & \multicolumn{1}{c|}{87.1\quad88.4} & \multicolumn{1}{c|}{88.1\quad88.4}  & \multicolumn{1}{c}{58.4\quad61.5} \\

10k    & \multicolumn{1}{c|}{TVIN}  & \multicolumn{1}{c|}{ \textbf{92.8\quad93.3}} & \multicolumn{1}{c|}{ \textbf{92.9\quad93.1}} & \multicolumn{1}{c|}{ \textbf{91.2\quad91.5}} & \multicolumn{1}{c|}{ \textbf{90.9\quad91.7}} & \multicolumn{1}{c|}{ \textbf{94.2\quad94.3}} & \multicolumn{1}{c|}{ \textbf{93.3\quad93.4}} & \multicolumn{1}{c|}{ \textbf{89.5\quad90.7}} & \multicolumn{1}{c|}{ \textbf{95.5\quad 95.5}} & \multicolumn{1}{c}{ \textbf{92.5\quad92.5}} \\
\bottomrule[1pt]
\end{tabular}}
\caption{Transfer from Drive to NEWS with varying dataset sizes N and maze sizes M.}
\end{table*}
%\vspace{-1em}
\paragraph{Accuracy w.r.t. domains} Based on these guidelines, we evaluate several instances of knowledge transfer, i.e., from NEWS to Moore, from Moore to NEWS and from Differential Drive to NEWS. For each transfer, we compare \tvin policy to the VIN reactive policy. Additionally we vary the maze sizes in each domains and dataset sizes in the target domains. Note that, $K$ is required to be chosen in proportion to the maze size. In the implementation, we refer to \cite{gppn2018} and set the default recurrence $K$ relative to the maze sizes: $K = 20$ for $9\times9$ mazes, $K = 30$ for $15\times15$ mazes and $K = 56$ for $28\times28$ mazes. Results are respectively reported in Table 1, Table 2, and Table 3, showing that our transfer learning approach \tvin provides a definite increase in accuracy when we have limited data in the target domain. Even compared to the standard reactive networks DQN of the success rate $74.2\%$ on Moore-28 with full dataset which is shown in \cite{DBLP:conf/nips/TamarLAWT16}, \tvin can reach the success rate of $84.3\%$ (Table 1), outperforming DQN only with 5k training data in the same case. Additionally, training process of the \tvin and \vin on 1k training data of Moore-15 is depicted in Figure 2. It also shows that knowledge transfer by \tvin speeds up learning process and reaches a higher generalization.
%\vspace{-1em}
\paragraph{Accuracy w.r.t. transfer methods}
As shown in Table 4, we make comparison with a simple transfer method denoted by VIN$_i$. VIN$_i$ is a heuristic transfer method \cite{Parisotto2015Actor} by directly leveraging pre-trained weights of $f_R$ and part of $f_p$ (with respect to similar actions) as the initialization for training in the target domain. Taking the experiments between NEWS-15 to MOORE for example, the results show that heuristic transfer by VIN$_i$ give useful pre-trained information, compared to training from scratch. Moreover, the \tvin policy learned in target domain performs much better than heuristic transfer VIN$_i$, which shows that our transfer strategies are effective and applicable.
%\vspace{-1em}
\paragraph{Accuracy w.r.t. planning complexity}
The complexity of planning in the 2-D maze domains generally depends on the number of obstacles and their distribution on the grid map. We thus synthesize domains based on different number of obstacles and different size of the grid map. In this experiments, We compare two complexity, which are 30 percent and 50 percent. It means 30 percent or 50 percent of the map is randomly placed with obstacles. Although we evaluate our approach on these 2-D domains, we should note that many real-world application domains, such as \emph{navigations}, \emph{warehouse scheduling}, etc. can be matched to 2-D maze domains with different complexity, and thus such evaluation in these domains should be convincing.

In this experiment, we view $9\times9$ NEWS as source domains, and transfer pre-trained knowledge to $15\times15$ Moore. We investigate the transfer performance with respect to different complexity. The results are show in Figure 2 and Figure 3, where the left one shows the transfer between domains of 30 percent obstacles, and the right one is the transfer between domains of 30 percent obstacles. In both cases, adjusting weights of the transferred knowledge in \tvin can indeed outperform the mechanism of randomly initializing VIN. It illustrates that \tvin planning policies, by our transfer strategies, are technically effective either in simple environment or complex environment. The performance gap between transfer learning policy \tvin and original VIN policy is more significant in low complexity domain, whereas in high complexity domains the gap between \tvin and VIN is comparatively slight. The difference in performance gap shows that it is more challenging for \tvin to leverage the pre-trained knowledge when the complexity of planning is much higher.
%Note that, to evaluate the generality of our transfer approach, we transfer the knowledge in \textbf{Grid-8-30} (with less complexity) to \textbf{Grid-16-50} (with more complexity), and \textbf{Grid-8-50} (with more obstacles) to \textbf{Grid-16-30} (with less obstacles). Transferring between identical domains (with identical size of maps and percentage of obstacles) is much easier than we do in this paper.
\begin{table}
	\resizebox{0.47\textwidth}{!}{
	\label{tab:actor}
	\centering
{
		\begin{tabular}{ccccc}
			\toprule[1.5pt]
			\multicolumn{2}{c|}{Source}                       & \multicolumn{3}{c}{NEWS-15} \\ \hline
			\multicolumn{2}{c|}{Target}                       & \multicolumn{1}{c|}{Moore-9}                  & \multicolumn{1}{c|}{Moore-15}                 & \multicolumn{1}{c}{Moore-28}
			  \\ \hline
			N                    & \multicolumn{1}{c|}{Model} & \multicolumn{1}{c|}{\%Opt    \%Suc}           & \multicolumn{1}{c|}{\%Opt    \%Suc}           & \multicolumn{1}{c}{\%Opt    \%Suc}            \\
			
			\midrule[0.5pt]
			
			1k                   & \multicolumn{1}{c|}{VIN}   & \multicolumn{1}{c|}{84.2\quad     87.7}          & \multicolumn{1}{c|}{77.3\quad     81.7}          & \multicolumn{1}{c}{56.2\quad     65.8}
			 \\
			1k                   & \multicolumn{1}{c|}{VIN$_i$}  & \multicolumn{1}{c|}{92.8\quad      94.9 }          & \multicolumn{1}{c|}{88.6\quad      91.2 }          & \multicolumn{1}{c}{65.2\quad     74.6 }
			\\
			1k                   & \multicolumn{1}{c|}{TVIN}  & \multicolumn{1}{c|}{\textbf{94.6\quad     96.6}} & \multicolumn{1}{c|}{\textbf{90.1\quad     92.8}} & \multicolumn{1}{c}{\textbf{66.1\quad     75.3}}
			\\
			\specialrule{0em}{1pt}{1pt}
            \hline
            \specialrule{0em}{1pt}{1pt}
			5k                   & \multicolumn{1}{c|}{VIN}   & \multicolumn{1}{c|}{90.5\quad     92.5}          & \multicolumn{1}{c|}{86.7\quad     88.7}          & \multicolumn{1}{c}{64.3\quad     72.9}
			 \\
			5k                   & \multicolumn{1}{c|}{VIN$_i$}  & \multicolumn{1}{c|}{96.2\quad      96.1 }          & \multicolumn{1}{c|}{94.2\quad      95.4 }          & \multicolumn{1}{c}{71.9\quad     80.9 }  \\
			5k                   & \multicolumn{1}{c|}{TVIN}  & \multicolumn{1}{c|}{\textbf{97.1\quad     97.2}} & \multicolumn{1}{c|}{\textbf{95.2\quad     96.0}} & \multicolumn{1}{c}{\textbf{73.4\quad     84.3}}
			\\
			\specialrule{0em}{1pt}{1pt}
            \hline
            \specialrule{0em}{1pt}{1pt}
			10k                  & \multicolumn{1}{c|}{VIN}   & \multicolumn{1}{c|}{86.2\quad     88.0}          & \multicolumn{1}{c|}{91.1\quad     92.3}          & \multicolumn{1}{c}{60.8\quad     68.0}          \\
			10k                  & \multicolumn{1}{c|}{VIN$_i$}  & \multicolumn{1}{c|}{96.1\quad      96.3 }          & \multicolumn{1}{c|}{95.0\quad      95.5 }          & \multicolumn{1}{c}{84.6\quad      90.4 }
			\\
			10k                  & \multicolumn{1}{c|}{TVIN}  & \multicolumn{1}{c|}{\textbf{97.4\quad     97.5}} & \multicolumn{1}{c|}{\textbf{96.2\quad     96.7}} & \multicolumn{1}{c}{\textbf{87.8\quad     91.8}}\\
			
			\bottomrule[1.5pt]
	\end{tabular}}}
	\caption{Policy performance compared with simple transferred VIN$_i$ and TVIN}
\end{table}
%\vspace{-1em}
\paragraph{Accuracy w.r.t. dataset sizes}
To evaluate the objective on transfer learning, we compare the performance of \tvin model by using different size of dataset. As is illustrated in Table 1, Table 2 and Table 3, the size of training data on target domain influences the performance of \tvin. Prediction accuracy with varying training data in target domain is also depicted in Figure 4. It shows that, in each case, \tvin can indeed outperform the mechanism of randomly initializing VIN. Although the performance gap decreases gradually with the dataset size increasing, the performance of \tvin turns out to be significantly greater than VIN when there is limited data in the target domain. This shows that if there is already sufficient data for a novel domain to learn optimal policies, information transferred from the source domain would not help improve the performance a lot. Rather, our transfer strategies focus more on generating planning-based \tvin policies for a target domain with limited dataset.
%\vspace{-1em}
\paragraph{Accuracy w.r.t. hyperparameters}
Following the above results that \tvin performs better or equals to VIN, we further evaluate the effect of varying both iteration count $K$ and kernel size $F$ on the \tvin models. Table 5 and Table 6 show $\%Opt$ and $\%Suc$ results of \tvin and VIN on Moore-15 for different values of F and K, and we use NEWS-9 as the source domain. This shows that \tvin outperforms VIN even when hyperparameters such as iteration count K and kernel size F are set differently in the target domains. Although in VINs, larger mazes require larger kernel sizes and iteration counts, the performance gap between \tvin and single VIN do not rely on a specific choice of hyperparameters.
%\vspace{-1em}
\paragraph{Accuracy w.r.t. transferred knowledge}
Finally, we evaluate the influence of the number of transferable actions between source and target domains in \tvin. The more actions are transferred, the more knowledge is leveraged in target domain. Table 7 shows results for different numbers of transferable actions between the source domain (NEWS-9) and the target domain (Moore-15) with 1k training data. It is illustrated that the more similar actions to transfer, the better performance for target \tvin to gain.
\begin{table}[t]
\centering

\begin{tabular}{l|l|l|l}
\toprule[1pt]
\multicolumn{1}{c|}{} & \multicolumn{1}{c|}{K = 10}         & \multicolumn{1}{c|}{K = 20}         & \multicolumn{1}{c}{K = 30}          \\
Model                 & \multicolumn{1}{c|}{\%Opt    \%Suc} & \multicolumn{1}{c|}{\%Opt    \%Suc} & \multicolumn{1}{c}{\%Opt     \%Suc} \\

\midrule[0.5pt]

VIN                   & \multicolumn{1}{c|}{70.3\quad      78.3}                      & \multicolumn{1}{c|}{67.7\quad      77.0}                      & \multicolumn{1}{c}{{64.7\quad      74.5}}                      \\
TVIN                  & \multicolumn{1}{c|}{\textbf{78.0\quad     85.8}}                       & \multicolumn{1}{c|}{\textbf{81.6\quad      90.8}}                      & \multicolumn{1}{c}{\textbf{80.1\quad      91.8}}                      \\
\bottomrule[1pt]
\end{tabular}
\caption{Test performance on Moore-15 transferred from NEWS-9 with varying iteration counts $K$.}
\end{table}
\begin{table}[t]
\centering
\begin{tabular}{l|l|l|l}
\toprule[1pt]
\multicolumn{1}{c|}{} & \multicolumn{1}{c|}{F = 3}          & \multicolumn{1}{c|}{F = 5}          & \multicolumn{1}{c}{F = 7}           \\
Model                 & \multicolumn{1}{c|}{\%Opt    \%Suc} & \multicolumn{1}{c|}{\%Opt    \%Suc} & \multicolumn{1}{c}{\%Opt     \%Suc} \\

\midrule[0.5pt]

VIN                   & \multicolumn{1}{c|}{64.7\quad    74.5}                        & \multicolumn{1}{c|}{77.3\quad    81.7}                        & \multicolumn{1}{c}{77.8\quad    83.1}                        \\
TVIN                  & \multicolumn{1}{c|}{\textbf{80.1\quad    91.8}}               & \multicolumn{1}{c|}{\textbf{88.3\quad    91.0}}               & \multicolumn{1}{c}{\textbf{85.3\quad    88.9}}               \\
\bottomrule[1pt]
\end{tabular}
\caption{Test performance on Moore15 transferred from NEWS-9 with varying kernel sizes $F$.}
\end{table}
 \section{Related Work}
In Reinforcement Learning (RL), the agent act in the world and learn a policy from trial and error. RL algorithms in \cite{Sutton2005Reinforcement,Schulman2015Trust,Levine2016End} use these observations to improve the value of the policy. Recent works investigate policy architectures that are specifically tailored for planning under uncertainty. VINs \cite{DBLP:conf/nips/TamarLAWT16} take a step in this direction by exploring better generalizing policy representations. The Predictron \cite{DBLP:conf/icml/SilverHHSGHDRRB17}, Value Prediction Network \cite{DBLP:conf/nips/OhSL17} also learn value functions end-to-end using an internal model, with recurrent neural networks (RNNs) \cite{DBLP:conf/interspeech/MikolovKBCK10} acting as the transition functions over abstract states. However, none of these abstract planning-based models have been considered for transfer. Our work investigates the generalization properties of the pre-trained policy and proposes the \tvin model for knowledge transfer.

\begin{table}[t]
\centering
\resizebox{0.47\textwidth}{!}{
\begin{tabular}{l|l|l|l|l}
\toprule[1pt]

\multicolumn{1}{c|}{Actions} & \multicolumn{1}{c|}{\begin{tabular}[c]{@{}c@{}}Num = 1\\ \%Opt     \%Suc\end{tabular}} & \multicolumn{1}{c|}{\begin{tabular}[c]{@{}c@{}}Num = 2\\ \%Opt    \%Suc\end{tabular}} & \multicolumn{1}{c|}{\begin{tabular}[c]{@{}c@{}}Num = 3\\ \%Opt    \%Suc\end{tabular}} & \multicolumn{1}{c}{\begin{tabular}[c]{@{}c@{}}Num = 4\\ \%Opt    \%Suc\end{tabular}} \\
\midrule[0.5pt]
VIN                                      & \multicolumn{1}{c|}{77.3\quad      81.7}                                                                         & \multicolumn{1}{c|}{77.3\quad      81.7}                                                                        & \multicolumn{1}{c|}{77.3\quad      81.7}                                                                        & \multicolumn{1}{c}{77.3\quad      81.7}                                                                       \\
TVIN                                     & \multicolumn{1}{c|}{82.0\quad      86.1}                                                                         & \multicolumn{1}{c|}{82.2\quad      86.5}                                                                        & \multicolumn{1}{c|}{86.2\quad      90.9}                                                                        & \multicolumn{1}{c}{\textbf{88.3\quad      91.0}}                                                              \\
\bottomrule[1pt]
\end{tabular}}
\caption{Test performance on Moore-15 transferred from NEWS-9 with varying number of transferred actions.}
\end{table}
A wide variety of methods have also been studied in the context of RL transfer learning\cite{Taylor2009Transfer}. Policy distillation \cite{Hinton2015Distilling,DBLP:conf/nips/ChenCYHC17} aims to compress the capacity of a deep network via efficient knowledge transfer . It has been successfully applied to deep reinforcement learning problems \cite{Rusu2016Policy}. Recently, successor
features and generalised policy improvement, has been introduced as a principled way of transferring skills \cite{DBLP:conf/icml/BarretoBQSSHMZM18}. Also \cite{DBLP:conf/icml/AbelJGKL18} considers value-function-based transfer in RL. However the key to our approach is that the Q-functions for specific actions learned from the source domain can be transferred to the corresponding VI module in the target domain. we also build a mapping between feature spaces in the source and target domains, transfer Q-networks related to \emph{similar actions} from the source to the target domain and build policy networks for \emph{dissimilar} actions which are learned from scratch.

 \section{Conclusions}
We propose a novel transfer learning approach \tvin to learn a planning-based policy for the target domain with different feature spaces and action spaces by leveraging pre-trained knowledge from source domains. In addition, we exhibit that such a transfer network \tvin leads to better performance when the training data is limited in the target domain.
In this paper we assume the pairs of possible similar actions is provided beforehand. In the future, it would be interesting to exactly learn the action similarities based on Web search \cite{DBLP:conf/aips/ZhuoYPL11,DBLP:journals/ai/Zhuo014} or language model learning \cite{DBLP:conf/atal/TianZK16,DBLP:conf/ijcai/FengZK18} before employing the transfer method.

\section*{Acknowledgement}
Hankz H. Zhuo thanks the support of the National Natural Science Foundation of China (U1611262), Guangdong Natural Science Funds for Distinguished Young Scholar (2017A030306028), Guangdong special branch plans young talent with scientific and technological innovation, Pearl River Science and Technology New Star of Guangzhou, Key Laboratory of Machine Intelligence and Advanced Computing (Sun Yat-Sen University) of Ministry of Education of China, and Guangdong Province Key Laboratory of Big Data Analysis and Processing for the support of this research. Sinno J. Pan thanks the support from NTU Nanyang Assistant Professorship (NAP) grant M4081532.020.

\bibliographystyle{aaai}
\bibliography{aaai}

\end{document}